\documentclass[10pt,twocolumn,letterpaper]{article}

\usepackage{cvpr}
\usepackage{times}
\usepackage{epsfig}
\usepackage{graphicx}
\usepackage{amsmath}
\usepackage{amssymb}
\usepackage{siunitx}
\usepackage{makecell}
\usepackage{array}
\usepackage{pdfpages}
\usepackage{wrapfig}
\usepackage{lipsum}
\usepackage{footnote}
\usepackage{multirow, dcolumn}
\usepackage{siunitx}
\usepackage{makecell}
\usepackage{epstopdf}
\usepackage{fancyhdr}
\usepackage{subcaption}
\usepackage[referable]{threeparttablex}
\usepackage[font=normalsize,skip=3pt]{caption}
\allowdisplaybreaks
\usepackage[font=small,labelfont=bf, skip=0pt]{caption}

\usepackage[pagebackref=true,breaklinks=true,letterpaper=true,colorlinks,bookmarks=false]{hyperref}
\usepackage{enumitem}
\setlist[enumerate]{noitemsep}
 \cvprfinalcopy 

\usepackage[normalem]{ulem}

\def\cvprPaperID{14} 

\makeatletter
\newcommand*{\rom}[1]{\expandafter\@slowromancap\romannumeral #1@}
\makeatother

\ifcvprfinal\fi
\begin{document}

\title{
{VGAN-Based} Image Representation Learning\\ for Privacy-Preserving Facial Expression Recognition
}
%

\author{Jiawei Chen\\
Boston University\\
{\tt\small garychen@bu.edu}
\and
Janusz Konrad\\
Boston University\\
{\tt\small jkonrad@bu.edu}
\and
Prakash Ishwar\\
Boston University\\
{\tt\small pi@bu.edu}
}

\maketitle
%
%

\begin{abstract}
	Reliable facial expression recognition plays a critical role in human-machine interactions. 
	However, most of the facial expression analysis methodologies proposed to date pay little or no attention to the protection of a user's privacy. 
	In this paper, we propose a Privacy-Preserving Representation-Learning Variational Generative Adversarial Network (PPRL-VGAN) to learn an image representation that is 
	explicitly disentangled from the identity information. 
	At the same time, this representation is discriminative from the standpoint of facial expression recognition and generative as it allows expression-equivalent face image synthesis. 
	We evaluate the proposed model on two public datasets under various threat scenarios. 
	Quantitative and qualitative results demonstrate that our approach strikes a balance between the preservation of privacy and data utility. 
	We further demonstrate that our model can be effectively applied to other tasks such as expression morphing and image completion. 
	\vglue -0.5cm
\end{abstract}

\section{Introduction}
The recent proliferation of sensors in living spaces is propelling the development of ``smart'' rooms that can sense and interact with occupants to deliver a number of benefits such as improvements in energy efficiency, health outcomes, and productivity~\cite{dai2015towards}. 
Automatic facial expression recognition is an important component of human-machine interaction.
To date, a wide variety of methods have been proposed to accomplish this, however they typically rely on high-resolution images and ignore the {\it visual privacy}~\cite{padilla2015visual} of users.
Growing privacy concerns will prove to be a major deterrent in the widespread adoption of camera-equipped smart rooms and the attainment of their concomitant benefits.
Therefore, reliable and accurate privacy-preserving methodologies for facial expression recognition are needed. 

\begin{figure}[!ht]
	\centering{\includegraphics[trim={0 0 0 0.6cm},clip, width=1.1\linewidth]{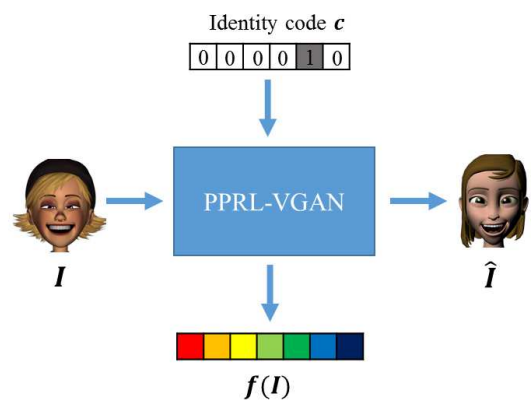}}
	\caption{
		Basic functionality of PPRL-VGAN: given an input face image $\boldsymbol{I}$, the
		network produces an identity-invariant representation $\boldsymbol{f}(\boldsymbol{I})$ , suitable for  facial expression recognition,  and an expression-preserving face image with another identity specified by identity code $\boldsymbol{c}$.}
	\label{fig:small_diagram}
	\vglue -0.6cm
\end{figure}

One approach to increase visual privacy is to reduce identity traits within a face image {\it via} modification or redaction methods such as pixelization or blurring. 
However, this will also reduce the visual quality of the modified image and an algorithm's ability to accurately recognize the facial expression from it.
Another extreme approach is to withhold releasing the face image altogether and only release an estimate of the facial expression. While this approach guarantees visual privacy, it provides no visual utility.
In order to strike a balance between privacy and data utility, we propose a third radically different approach: seamlessly {\it replace} the user-identity in an image without significantly degrading its visual quality or the ability to accurately infer facial expression. We leverage variational generative-adversarial networks (VGANs) to learn an identity-invariant representation of an image while enabling the synthesis of a utility-equivalent, realistic version of this image with a different identity (Fig.~\ref{fig:small_diagram}). We call this framework Privacy-Preserving Representation-Learning Variational Generative Adversarial Network (PPRL-VGAN). Beyond its application to privacy-preserving visual analytics, our approach could also be used to generate realistic avatars for animation and gaming. 

Our proposed framework combines the generative power of two models: the Variational Auto-Encoder (VAE)~\cite{kingma2013auto} and the Generative Adversarial Network (GAN)~\cite{goodfellow2014generative}. A VAE consists of two networks: the encoder,  which maps a data sample to a latent representation,  and the decoder,  which maps this representation back to data space. 
VAE networks are trained by minimizing a cost function that encourages learning a latent representation which leads to realistic data synthesis while ensuring sufficient diversity in the synthesized data.
%
Like a VAE, a GAN also consists of two networks: a generator network ($G$) which aims to synthesize realistic data from a random noise input vector and a discriminator network ($D$) which aims to differentiate between real and synthetic data. GANs are trained \textit{via} a game between $G$ and $D$ in which $G$ aims to fool $D$ into believing that the data samples synthesized by it are realistic, and $D$ which aims to accurately distinguish between real and ``fake'' samples.
%
In this work, we combine VAEs with GANs by replacing the generator in a conventional GAN, which uses random noise as input, with a VAE encoder-decoder pair, which takes a real image as an input and outputs a synthesized image.
%
As shown in Fig.~\ref{fig:proposed network}, the encoder learns a mapping from a face image $\boldsymbol{I}$ to a latent representation $\boldsymbol{f}(\boldsymbol{I})$.
The representation is subsequently fed into the decoder to synthesize a face image with some target identity (specified by identity code $\boldsymbol{c}$) but with the same facial expression as the input image.
The discriminator includes multiple classifiers that are trained to (i) distinguish real face images from synthesized ones, (ii) recognize the identity of the person in a face image and (iii) recognize the expression in a face image.   
During training, feedback signals from $D$ guide $G$ to create realistic expression-preserving face images. In addition, as the identity of the synthesized images is determined by the identity code $\boldsymbol{c}$, the network will learn to disentangle the identity-related information from the latent representation. 

This paper makes the following contributions:
\begin{enumerate}[topsep=0ex]
	\item We propose a framework for learning an identity-invariant representation for a face image. This representation is discriminative for facial expression recognition and generative for expression-preserving, identity-altered face image synthesis.
	\item We thoroughly evaluate our approach under three threat scenarios to demonstrate that our method strikes a balance between privacy and data utility. 
	\item We demonstrate that our model can synthesize new face images with or without an input image, and illustrate how our model can also be applied to other image processing tasks such as expression morphing and image completion. 
\end{enumerate} 

\section{Related Work}
\textbf{Privacy-Preserving Visual Analytics:} There is a growing body of research
on methods to perform various visual analysis tasks from data in a manner that does not disclose subject's identity. 
According to how privacy is protected, the literature can be broadly classified as reversible and
irreversible approaches~\cite{badii2013holistic}.

{\it Reversible methods} include scrambling and encryption~\cite{dufaux2006scrambling,sadeghi2009efficient,wang2017encrypted, ziad2016cryptoimg} that permit exact data recovery, but are also prone to exposing the original data to possible hacks.
%
%
In particular, methods for recognizing facial expression directly in the encrypted domain have been proposed \cite{aina2014spontaneous,rahulamathavan2013facial}. However, these methods rely upon public-key homomorphic cryptosystems, such as Paillier~\cite{paillier1999public}, which are known to be computationally heavy due to their use of large encryption and decryption keys.
In order to relieve the computational burden, lightweight algorithms based on randomization techniques have been proposed in \cite{rahulamathavan2017efficient}. 
Although  methods proposed in~\cite{aina2014spontaneous,rahulamathavan2013facial, rahulamathavan2017efficient}
perform well for facial expression recognition in the encrypted domain, 
no tests have been conducted to ascertain whether the identity information is indeed removed in the encrypted  domain. It is unclear whether a classifier that is trained on encrypted-domain images will fail to recognize the identity of a person from the encrypted image.

{\it Irreversible methods} include image processing and filtering techniques~\cite{chen2017semi, chen2016estimating,dai2015towards,jalal2012depth,krinidis2014robust,park2008privacy,roeper2016privacy}. 
However, it has been shown that simple filtering methods do not fool identity-recognition algorithms if they are trained using images that have the same distortion as the test images~\cite{newton2005preserving}.
%
A face de-identification method was proposed in~\cite{jourabloo2015attribute} wherein several face images with appearance attributes similar to the target image are fused by minimizing a cost function promoting attribute preservation and de-identification.
A recent line of irreversible methods makes use of adversarial networks~\cite{brkic2017know, pittaluga2018learning,raval2017protecting}. In~~\cite{brkic2017know}, the focus is on full-body de-identification without an additional utility criterion such as accuracy of facial expression. 
Their methodology also relies upon a segmentation algorithm to accurately extract the silhouette of the person to be de-identified. Moreover, the synthesized images are blurry. 
While \cite{raval2017protecting} uses adversarial networks to jointly optimize privacy and utility objectives, it focuses on the relatively simple task of detecting and removing a QR code embedded in an image. Moreover, the synthesized images are poor-quality renderings of the input image.
The approach in \cite{pittaluga2018learning} is similar in spirit to \cite{raval2017protecting} but the output is not required to look realistic.
Our approach differs from these methods in that we use a VAE within a GAN in order to explicitly learn an identity-invariant facial expression representation with the explicit goal of expression-preserving identity replacement in the synthesized output image which is required to look realistic.
%
As we show, our learned representation is not only discriminative for expression recognition, but also robust to both human and algorithm-based privacy attacks. Our framework can also be used for other tasks such as expression morphing. 

\textbf{Disentangled Representation Learning:}
A number of models have been proposed in the literature to learn a so-called ``disentangled representation''. 
In early work,  a bilinear model was proposed to separate content and style for face and text images~\cite{tenenbaum1997separating}. 
An autoencoder (AE) augmented with simple regularization terms during training was proposed in~\cite{cheung2014discovering} and demonstrated to discover and explicitly learn various latent factors of variation. 
Methods proposed in ~\cite{kingma2014semi,makhzani2015adversarial}  use VAEs in a semi-supervised manner. Their models disentangle label information from the latent representation by providing additional labels as input to the decoder. 
However, methods based on AE/VAE tend to produce blurry images due to the pixel-wise reconstruction error used in the loss function.
Our model may be viewed as replacing the image reconstruction error with an adversarial loss to improve the visual quality of synthesized images. 
%
Recently, a two-stage pipeline was proposed~\cite{ma2017disentangled} to learn disentangled image representations of background, foreground, and pose to generate novel person images. However, this method requires a pre-processing step to estimate a coarse pose mask of the input image.

Among works on disentangled representation learning, perhaps the closest to ours are those in ~\cite{mathieu2016disentangling,tran2017disentangled}.
The approach proposed in ~\cite{mathieu2016disentangling} addresses the problem of disentaglement by combining a deep convolutional VAE with a form of adversarial training.
It can disentangle the latent factors of variation within a labeled dataset, and separate them into complementary codes.
However, it has not been tested on a real-world dataset. 
Our approach is different from that in~\cite{mathieu2016disentangling} as we completely discard the VAE's reconstruction error in the objective function. Instead, we employ the adversarial loss from a GAN for high-quality image synthesis and improved representation learning. 
In \cite{tran2017disentangled},  a disentangled representation-learning GAN was proposed for pose-invariant face recognition. 
The proposed model is a fusion of an AE and a GAN. It explicitly disentangles the identity representation from pose variation by passing a pose code to the decoder during training. 
The major difference between this model and ours is that in PPRL-VGAN we use a VAE instead of an AE which permits learning a probability distribution over the latent space.
This enables our model to synthesize new images without an input image; all we need to do is generate a latent vector from the prior distribution and pass it to the decoder along with an identity code. 

\section{Background Material}
\begin{figure*}[!ht]
	\centering{\includegraphics[trim={0 0 0 0.6cm},clip, width=0.8\linewidth]{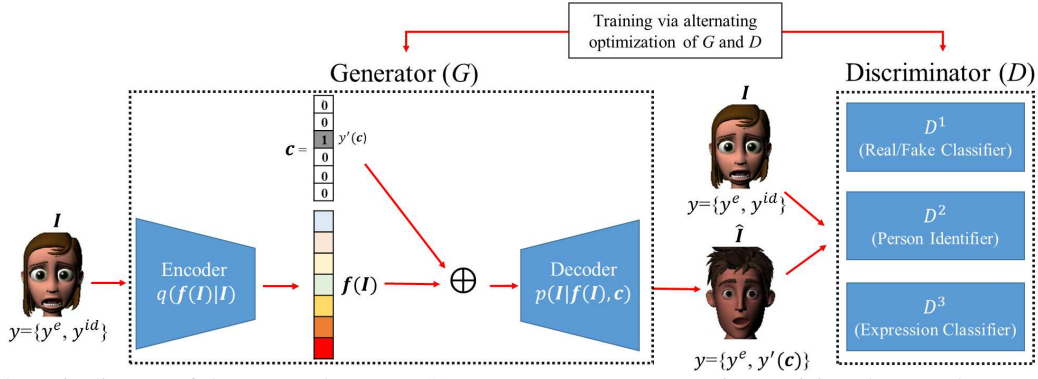}}
	\caption{Schematic diagram of the proposed PPRL-VGAN ($\oplus$ represents concatenation).{Training alternates between optimizing the weights of $D$ keeping $G$ fixed and vice-versa. Both original and synthesized images with their labels are used during training.}
	}
	\label{fig:proposed network}
	\vglue -0.5cm
\end{figure*}
\subsection{Variational Autoencoder Network}
A VAE network consists of two neural networks: an encoder network ($Enc$) and a decoder network ($Dec$).  The encoder is a randomized mapping of a data sample $\boldsymbol{x}$ to a latent representation $\boldsymbol{z}$ while the decoder is a randomized mapping $\boldsymbol{z}$ from a latent representation back to data space:
\begin{equation}
\boldsymbol{z} \sim Enc(\boldsymbol{x}) = q(\boldsymbol{z}|\boldsymbol{x})
\end{equation}
\begin{equation}
\boldsymbol{\widehat{x}} \sim Dec(\boldsymbol{z}) = p(\boldsymbol{x}|\boldsymbol{z})
\end{equation}
In practice, these randomized mappings are implemented {\it via} deterministic maps (given by the neural networks) with additional inputs which provide the source of randomness.
For example, it is common to set
$
\boldsymbol{z} = \boldsymbol{\mu}_{\boldsymbol{x}} + \boldsymbol{A}_{\boldsymbol{x}}\boldsymbol{w}
$
where the vector $\boldsymbol{\mu}_{\boldsymbol{x}}$ and the square matrix $\boldsymbol{A}_{\boldsymbol{x}}$ are the outputs of a neural network with input $\boldsymbol{x}$, and $\boldsymbol{w} \sim \mathcal{N}(\boldsymbol{0},\boldsymbol{I})$, a standard multivariate Gaussian, is the source of randomness. Then, $q(\boldsymbol{z}|\boldsymbol{x}) = \mathcal{N}(\boldsymbol{\mu}_{\boldsymbol{x}},\boldsymbol{A}_{\boldsymbol{x}}\boldsymbol{A}_{\boldsymbol{x}}^T)$.
VAE networks are trained by {\it minimizing} a cost function which is additive over all training data samples. The cost function for a single data sample $\boldsymbol{x}$ is given by 
\begin{equation}
\label{eq:vae}
\mkern-12mu \mathcal{L}^{VAE}_{\boldsymbol{x}}= -\mathbb{E}_{\boldsymbol{z}\sim q(\boldsymbol{z}|\boldsymbol{x})} [\log p(\boldsymbol{x}|\boldsymbol{z}) ] + KL\big(q(\boldsymbol{z}|\boldsymbol{x}) || p(\boldsymbol{z}) \big)
\end{equation}
where $KL$ is the Kullback-Leibler divergence and $p(\boldsymbol{z})$, the marginal distribution of the latent representation, is typically taken to be 
$\mathcal{N}(\boldsymbol{0},\boldsymbol{I})$.
The first term encourages the decoder to assign higher probability to the observed data samples $\boldsymbol{x}$. In practice, the expectation in the first term is replaced by an empirical average across a small batch of independent and identically distributed $\boldsymbol{z}$ for a given $\boldsymbol{x}$.
The $KL$ term encourages the encoder $q(\boldsymbol{z}|\boldsymbol{x})$ to be close to a target $p(\boldsymbol{z})$ which has sufficient spread (diversity) in the latent space. The $KL$ term has a closed analytic form since both its arguments are Gaussian~\cite{kingma2013auto}. The total cost across all data samples is typically minimized {\it via} {mini-batch gradient descent}.

\subsection{Generative Adversarial Network }
A standard GAN consists of a generator neural network $G$ and a discriminator neural network $D$ that are trained by making them compete in a two-player min-max game. 
The discriminator network \textit{D} adjusts its weights so as to reliably distinguish real data samples $\boldsymbol{x} \sim p_{d}(\boldsymbol{x})$ from fake data samples $G(\boldsymbol{z})$ generated by passing $\boldsymbol{z}$, randomly sampled from some distribution $p_{z}(\boldsymbol{z})$, through the generator network $G$. The generator network $G$ adjusts its weights to fool $D$.
%
%
The discriminator \textit{D} assigns probability $D(\boldsymbol{x}) \in [0,1]$ to the event that $\boldsymbol{x}$ is a ``real'' training data sample and the probability $1-D(\boldsymbol{x})$ to the event that $\boldsymbol{x}$ is a ``fake'' sample synthesized by the generator. 
The two networks are trained iteratively using a loss function given by
%
\begin{align}
\mathcal{L}_{GAN}(G,D) =&\ E_{\boldsymbol{x} \sim p_d(\boldsymbol{x})}[\log D(\boldsymbol{x})]\ + \nonumber \\
%
%
&\ E_{\boldsymbol{z} \sim p_{z}(\boldsymbol{z})}[\log(1 - D(G(\boldsymbol{z}))]
\end{align}
%
with $G$ aiming to minimize $\mathcal{L}_{GAN}(G,D)$ and $D$ aiming to maximize it. In practice, the expectations are replaced by empirical averages over a mini-batch of samples and the loss function is alternately minimized and maximized from one mini-batch to the next as in {mini-batch gradient descent}.

\section{Formulation of PPRL-VGAN}
%
%
%
%
Given a face image $\boldsymbol{I}$ with an identity label $y^{id} = 1,...,N_{id}$ and an expression label $y^{e} = 1,...,N_e$, where $N_{id}$ and $N_e$ are the numbers of distinct subjects and facial expressions, respectively,
the proposed model has two objectives:
1) to learn an identity-invariant face image representation $\boldsymbol{f}(\boldsymbol{I})$ for facial expression recognition, and
2) to synthesize a realistic face image ${\boldsymbol{\widehat{I}}}$ with the same facial expression as in $\boldsymbol{I}$ and target identity specified by a {one-hot encoded identity code $\boldsymbol{c} \in \{0,1\}^{N_{id}}$.} 

\noindent{\bf Discriminator:} Different from the discriminator network in a conventional GAN, the discriminator $D$ $=$ $(D^1,$ $D^2,$ $D^3)$ in PPRL-VGAN is a multi-task classifier consisting of three separate neural networks (Fig.~\ref{fig:proposed network}): 1) the $D^{1}$ network classifies an input face image $\boldsymbol{I}$ as real or synthetic, 2) the $D^2$ network estimates the identity of the person in the input face image, and 3) the $D^3$ network classifies the facial expression in the input face image. 
The weights of the networks in $D$ are trained to classify real face image inputs $\boldsymbol{I}$ as real and accurately recognize the person's identity and the facial expression. They are also trained to classify  synthetic image inputs  $\boldsymbol{\widehat{I}}$ as fake. This is accomplished by adjusting the network weights to {\it maximize} the following {\it discriminator cost function}:
%
%
\begin{align}
\mathcal{L}_{D}&(G,D) = 
\lambda^{D}_{1} \left\{E_{\boldsymbol{I} \sim p_d(\boldsymbol{I})}[ \log D^{1}(\boldsymbol{I})]\right. + \nonumber \\
& \left.E_{\boldsymbol{I} \sim p_d(\boldsymbol{I}), \boldsymbol{c}\sim p(\boldsymbol{c})}[\log(1 - D^{1}(G(\boldsymbol{I}, \boldsymbol{c})))] 
\right\} + \nonumber \\
&E_{(\boldsymbol{I},\boldsymbol{y}) \sim p_d(\boldsymbol{I}, \boldsymbol{y})}[
\lambda^{D}_2 \log D^{2}_{y^{id}}(\boldsymbol{I}) + \lambda^{D}_3 \log D^{3}_{y^{e}}(\boldsymbol{I}) ]
\label{eq:pprlvgan_disc}
\end{align}
where $D^2_{i}$, $D^3_{i}$ are the predicted probabilities of the $i$th class for identity and facial expression, respectively. 
The tuning parameters $\lambda^{D}_1$, $\lambda^{D}_2$ and $\lambda^{D}_3$ control the relative importance between image quality, identity recognition, and expression recognition objectives. 

\noindent{\bf Generator:} In contrast to the generator in a conventional GAN which directly maps a ``noise'' vector to a synthesized image, the generator $G$ in a PPRL-VGAN maps a real input image $\boldsymbol{I}$ with identity $y^{id}$ and expression $y^e$ to a synthesized output image $\boldsymbol{\widehat{I}} = G(\boldsymbol{I}, \boldsymbol{c})$ with a target identity $y'(\boldsymbol{c})$ and the same expression $y^e$. This is accomplished {\it via} a VAE-like encoder-decoder structure.
Specifically, the encoder aims to learn an image representation $\boldsymbol{f}(\boldsymbol{I})$ from $\boldsymbol{I}$ {\it via} a randomized mapping $\boldsymbol{f}(\boldsymbol{I}) \sim q(\boldsymbol{f}(\boldsymbol{I})|\boldsymbol{I})$ parameterized by the weights of the encoder neural network. 
Similarly to a VAE, the cost function for training the generator includes $KL$ divergence between  a prior distribution on the latent space $p(\boldsymbol{f}(\boldsymbol{I})) \sim \mathcal{N}(\boldsymbol{0},\boldsymbol{I})$ and the conditional distribution $q(\boldsymbol{f}(\boldsymbol{I})|\boldsymbol{I}))$. Training attempts to minimize this $KL$ term.
The generator cost function also includes a term that encourages the decoder to learn to synthesize a face image ${\boldsymbol{\widehat{I}}} \sim p(\boldsymbol{I}|\boldsymbol{f}(\boldsymbol{I}),\boldsymbol{c})$ that can fool
$D$ into classifying it as a real face image having the same facial expression $y^e$ as the input image $\boldsymbol{I}$, but with a target identity $y'(\boldsymbol{c})$ determined by $\boldsymbol{c}$.
Specifically, the generator network weights are adjusted during training to {\it minimize} the following {\it generator cost function}:
\begin{align}
&\mathcal{L}_{G}(G,D) = \nonumber \\
& E_{(\boldsymbol{I},\boldsymbol{y}) \sim p_d(\boldsymbol{I},\boldsymbol{y}), \boldsymbol{c}\sim p(\boldsymbol{c})}[
\lambda^{G}_1 \log (1-D^{1}(G(\boldsymbol{I}, \boldsymbol{c}))) + \nonumber \\
& \lambda^{G}_2 \log (1-D^{2}_{y'(\boldsymbol{c})}(G(\boldsymbol{I}, \boldsymbol{c}))) + \lambda^{G}_3 \log (1 -D^{3}_{y^{e}}(G(\boldsymbol{I}, \boldsymbol{c}))) ] \nonumber \\
& + \lambda^{G}_4 KL\big(q(\boldsymbol{f}(\boldsymbol{I})|\boldsymbol{I}) || p(\boldsymbol{f}(\boldsymbol{I})) \big)
\label{eq:pprlvgan_gen}
\end{align}

\noindent
where $\lambda^{G}_1$, $\lambda^{G}_2$, $\lambda^{G}_3$ and $\lambda^{G}_4$ are tuning parameters of the loss functions for $D^1$, $D^2$, $D^3$ and $KL$ divergence respectively. 
A key difference compared to the cost in Eq.~\ref{eq:vae} is that first term (reconstruction error) in Eq.~\ref{eq:vae} has been replaced with a perceptual loss term for the discriminator $D^1$ in Eq.~\ref{eq:pprlvgan_gen}. 

Training alternates between maximizing Eq.~\ref{eq:pprlvgan_disc} with respect to the weights of the networks in $D$ and minimizing Eq.~\ref{eq:pprlvgan_gen} with respect to the weights of the networks in $G$. 
As the target identity code $\boldsymbol{c}$ ranges over all $N_{id}$ distinct subjects, $N_{id}$ synthetic images $\widehat{\boldsymbol{I}}$ are produced for each training or test image $\boldsymbol{I}$.
As in the training of VAEs and GANs, the expectations are approximated by empirical averages computed from a mini-batch of training examples.
Over successive training epochs, $G$ learns to fit the true data distribution and create a realistic face image that can fool $D^1$ having the same facial expression as the input image, which can be correctly recognized by $D^3$, and identity $y'(\boldsymbol{c})$, which can be correctly recognized by $D^2$.
As the latent code $\boldsymbol{c}$ determines the identity of ${\boldsymbol{\widehat{I}}}$, the encoder is encouraged to disentangle the identity information from $\boldsymbol{f}(\boldsymbol{I})$. 
Moreover, as ${\boldsymbol{\widehat{I}}}$ retains information about facial expression, the encoder is also encouraged to embed as many expression attributes as possible into $\boldsymbol{f}(\boldsymbol{I})$.
As a consequence, $\boldsymbol{f}(\boldsymbol{I})$ is a generative representation that is not only invariant to identity, but also discriminative for facial expression recognition.

\section{Experimental Evaluation}
\subsection{Datasets}
In order to validate the effectiveness of the proposed model, we conducted experiments on two public facial expression datasets: FERG~\cite{aneja2016modeling} and MUG~\cite{aifanti2010mug}.
%
FERG is a database of cartoon characters with annotated facial expressions containing 55,769 annotated face images of six characters.
The images for each character are grouped into 7 types of cardinal expressions, {\it viz.} anger, disgust, fear, joy, neutral, sadness and surprise.
%
%
The MUG database is video-based. It consists of realistic image sequences of 86 subjects performing the same 7 cardinal expressions. 
For the sake of computational efficiency, we chose the 8 subjects having the most image samples as our training and testing data. 
In each image sequence, we removed the first and last 20 frames which mostly correspond to the neutral expression. 
We used 11,549 images in total. 
In experiments with both datasets, we randomly selected (without replacement) $85\%$ images of each expression from each subject for the training set. 
The remaining $15\%$ of images were used as testing data. 
We also resized each RGB image to $64\times64$-pixel resolution.

\subsection{Training Details}
We used the same network architecture for both datasets. Details of  PPRL-VGAN structure are listed in Table~\ref{tbl:architecture}.
We implemented our algorithm in Keras~\cite{chollet2015} and trained all networks from scratch.
%
The weights were initialized to be zero-mean Gaussian with a small standard deviation of $10^{-2}$.
We used a batch size of 256 and performed batch normalization after each convolutional/deconvolutional
layer except the last deconvolutional layer in the decoder.
We set $ \alpha= 0.2$ for LeakyReLU's across the network. 
We used RMSprop optimizer~\cite{hinton2012rmsprop} with a learning rate of $0.0002$.
We observed that  network training is very sensitive to the choice of the tuning parameters in the generator and discriminator cost functions. 
We optimized these parameters using grid search. 
We found that the following values:  $\lambda^{D}_1 = 0.25$, $\lambda^{D}_2 = 0.5$, $\lambda^{D}_3 = 0.25$ for discriminator training and $\lambda^{G}_1 = 0.108$, $\lambda^{G}_2 = 0.6$, $\lambda^{G}_3 = 0.29$ , $\lambda^{G}_4 = 0.002$ for generator training work well. 
In conventional GANs, it is common to optimize the discriminator more frequently than the generator. 
However, we update the generator twice as frequently as the discriminator in training because the class labels used in PPRL-VGAN provide additional labeled data that help the discriminator training.
%
%
\begin{table*}[!htb]\small
	\caption{Architecture of PPRL-VGAN. $\downarrow$ and $\uparrow$ represent down- and upsampling operations,  respectively. $D^1$, $D^2$ and $D^3$ share the weights of all convolutional layers and of the first fully-connected layer. }
	\medskip
	\centering
	\resizebox{\textwidth}{!}{
		\begin{threeparttable}
			\begin{tabular}{c l l l}
				\Xhline{1pt}
				\multicolumn{1}{c}{Layer}& \multicolumn{1}{c}{Encoder} &  \multicolumn{1}{c}{Decoder} & \multicolumn{1}{c}{Discriminator}   \\
				\hline
				1 & $5\times5\times32$ conv. $\downarrow$, BNorm, LeakyReLU& 2048 FC layers $\xrightarrow{\text{Reshape}}$ $4\times4\times128$ , LeakyReLU & 	$5\times5\times32$ conv, BNorm, LeakyReLU\\
				2 &$5\times5\times64$ conv. $\downarrow$, BNorm, LeakyReLU& $5\times5\times256$ deconv. $\uparrow$, BNorm, LeakyReLU  & $5\times5\times64$ conv, BNorm, LeakyReLU \\
				3 &$5\times5\times128$ conv. $\downarrow$, BNorm, LeakyReLU& $5\times5\times128$ deconv. $\uparrow$, BNorm, LeakyReLU & $5\times5\times128$ conv, BNorm, LeakyReLU \\
				4 &$5\times5\times256$ conv. $\downarrow$, BNorm, LeakyReLU& $5\times5\times64$ deconv. $\uparrow$, BNorm, LeakyReLU & $5\times5\times256$ conv, BNorm, LeakyReLU\\
				5 & 128 fully-connected (FC), Linear & $5\times5\times3$ deconv, tanh& 256 fully-connected, LeakyReLU\\
				6 & & & $D^1$: 1 FC , $D^2$: $N_{id}$ FC , $D^3$: $N_e$ FC \\
				\Xhline{1pt}
			\end{tabular}
		\end{threeparttable}}
		\vglue -0.5cm
		\label{tbl:architecture}
	\end{table*}
	The source code, additional implementation details and more experimental results are available on our project website~\cite{website}. 
	
	\subsection{Threat Scenarios}
	\label{TS}
	We evaluate privacy-preserving performance of the proposed PPRL-VGAN under three threat scenarios. 
	
	\noindent{\bf Attack scenario \rom{1}}: This is a simple scenario in which the attacker has access to the unaltered training set $(\boldsymbol{I}_{train}, y_{train}^{id})$. However, {the attacker's test set consists of} all images in the {original test set} {\it after} they have been passed through the trained PPRL-VGAN network. Thus, the attacker never gets to see the original test image $\boldsymbol{I}_{test}$ but only its privacy-protected version $\widehat{\boldsymbol{I}}_{test}$. Also, the test set for the attacker contains all $N_{id}$ distinct privacy-protected versions $\widehat{\boldsymbol{I}}_{test}$ of each $\boldsymbol{I}_{test}$ corresponding to $N_{id}$ distinct values of the identity code $\boldsymbol{c}$.

	\noindent{\bf Attack scenario \rom{2}}: This is a more challenging scenario (from the perspective of protecting privacy) where the attacker has access to the privacy-protected training images ${\boldsymbol{\widehat{I}}}_{train}$ and knows their underlying ground-truth identities $y_{train}^{id}$. 
	Therefore, the attacker can train an identifier on training images that have the same type of identity-protecting transformation as the test images.
	If the proposed privacy-preserving transformation is weak and the identifier has sufficient learning capacity, it may be possible for a trained identifier to correctly predict the underlying ground-truth identity even from a privacy-protected test image.
	Similarly to scenario \rom{1}, there are $N_{id}$ images for each training and testing image.
	%
	
	\noindent{\bf Attack scenario \rom{3}}: In this scenario, the attacker gets access to the encoder network and can obtain the latent representation $\boldsymbol{f}(\boldsymbol{I})$ for any image $\boldsymbol{I}$.
	Then, if the produced latent representation is not void of identity traits, the attacker can train an identifier using $(\boldsymbol{f}(\boldsymbol{I}_{train}), y_{train}^{id})$ and apply it to $\boldsymbol{f}(\boldsymbol{I}_{test})$ for identification. Although more challenging than scenario \rom{2}, because the attacker can access the ``more pristine'' $\boldsymbol{f}$, there are fewer training and test samples available since the identity code $\boldsymbol{c}$ does not enter into the picture and thus there is no $N_{id}$-fold dataset expansion. Moreover whereas $\widehat{\boldsymbol{I}}$ resembles a real image, $\boldsymbol{f}$ needs not (and typically does not). 

	In terms of utility, we train a dedicated facial expression classifier in each scenario with the available format of training data and the corresponding ground-truth expression labels. 
	Then, we apply this classifier to test data and measure the facial expression recognition performance.
	
	\subsection{Privacy Preservation versus Data Utility}
	We first conduct a detailed evaluation of the proposed framework with respect to privacy preservation and data utility.  
	We use correct classification rate (CCR) in person identification to measure  how much privacy is preserved (the lower the CCR, the better) and also in facial expression recognition to measure the  utility of data (the higher the CCR, the better).
	Table~\ref{tbl:FERG_results}  summarizes the performance of the proposed approach on the FERG and MUG datasets under a privacy-unconstrained scenario (training and testing sets are both unaltered), under a random-guessing attack and under the three attack scenarios described earlier.
	In each scenario, the identification and facial expression are estimated separately by different neural network classifiers.

	\begin{table}[!thb]
		\vglue -0.30cm
		\caption{Person identification and facial expression recognition performance in different scenarios on FERG and MUG datasets.  }
		\label{tbl:FERG_results}
		\centering
		\resizebox{\columnwidth}{!}{%
			\begin{tabular}{ c c c c c }
				\Xhline{1pt}
				\multirow{2}{*}{ Scenario}	
				&   \multicolumn{2}{c}{Identification} 
				&   \multicolumn{2}{c}{Expression Recognition} 
				\\\cline{2-5}
				& FERG & MUG & FERG & MUG \\
				\hline			
				Privacy Unconstrained & $100\%$ & $100\%$ & $100\%$ & $87.90\%$\\
				\hline
				Random Guess & $16.67\%$ & $12.50\%$ & $14.29\%$ & $14.29\%$ \\
				\hline
				Attack Scenario \rom{1} & $17.01\%$ & $12.80\%$ & $93.02\%$ & $82.33\%$\\
				\hline 
				Attack Scenario \rom{2} & $28.30\%$ & $22.08\%$ & $95.00\%$ & $85.14\%$\\
				\hline 
				Attack Scenario \rom{3} & $22.42\%$ & $20.62\%$ & $100.00\%$ & $87.58\%$\\
				\Xhline{1pt}
			\end{tabular}}
			\vglue -0.35cm
		\end{table}
		For attack scenario \rom{1}, we train an identifier using the original training set $(\boldsymbol{I}_{train}, y_{train}^{id})$ and apply it to privacy-protected test images ${\boldsymbol{\widehat{I}}}_{test}$. The identifier has the same structure as $D^2$ (Fig.~\ref{fig:proposed network}).
		We first observe that the identification CCRs are $17.01\%$ for FERG and $12.80\%$ for MUG. Both are llose to a random guess ($16.67\%$ for FERG since there are 6 characters and $12.50\%$ for MUG since we selected 8 subjects).
		However, the same classifier applied to the privacy-unconstrained test images results in  $100\%$ identification performance on both datasets.
		Such a huge performance gap confirms the proposed model effectively protects users' privacy when the attacker has no information about the applied privacy-preserving transformation.
		For utility evaluation, we train a dedicated facial expression classifier, with the same structure as $D^3$,  using $(\boldsymbol{I}_{train}, y_{train}^{e})$ pairs and test it on ${\boldsymbol{\widehat{I}}}_{test}$ images.
		The resulting expression recognition accuracies are $93.02\%$ for FERG and $82.33\%$ for MUG. 
		These results are close to those achieved in the privacy-unconstrained scenario, which indicates that the synthesized images look realistic and retain the expression of the input images. 
		
		In  attack scenario \rom{2}, we use the privacy protected training data ${\boldsymbol{\widehat{I}}}_{train}$ and the corresponding ground-truth identity labels to train an identity recognizer and the ground-truth expressions to train a facial expression classifier (having the same architectures as in scenario \rom{1}).
		%
		%
		We first observe that the identification accuracy in scenario \rom{2} is
		about $11\%$ higher than that of a random guess
		for both datasets, which suggests that some identity-related information is leaked into the synthesized images, but this
		is still much lower than in the privacy-unconstrained scenario.
		With respect to facial expression recognition,  the performance in scenario \rom{2} is consistently better  than that in scenario \rom{1}. 
		This is likely because the number of training samples in scenario \rom{2} is $N_{id}$ times that in scenario \rom{1}, which benefits the training of the facial expression classifier.
		
		In attack scenario \rom{3}, we assume the attacker can access the latent representations of the training and probe images. 
		We simulate this attack scenario by training an identifier using $(\boldsymbol{f}(\boldsymbol{I}_{train}), y_{train}^{id})$ and test it on $\boldsymbol{f}(\boldsymbol{I}_{test})$.
		However, as $\boldsymbol{f}(\boldsymbol{I})$ is a 1-D vector, the 2-D ConvNet classifiers we used before are not suitable. 
		We have experimented with 3 classifiers for $\boldsymbol{f}(\boldsymbol{I})$, namely  a Support Vector Machine (SVM), a customized 1-D ConvNet and a customized Artificial Neural Network (ANN). 
		The customized ANN (3 hidden layers, each with 256 nodes) performed best in terms of identification and expression recognition accuracy. 
		Therefore, only results for the customized ANN classifier are reported.
		As shown in Table \ref{tbl:FERG_results}, the identification performance is reduced in comparison with scenario \rom{2}. However, the expression recognition performance in scenario \rom{3} is the best among the three attack scenarios.
		%
		Effectively, this suggests that the learned image representation $\boldsymbol{f}(\boldsymbol{I})$ contains crucial facial expression information,  
		but is largely disentangled from the identity information.

		\noindent{\bf Identity Replacement/Expression Transfer:}
		In addition to producing an identity-invariant image representation, PPRL-VGAN can be applied to an input face image of any identity to synthesize a realistic, expression-equivalent output face image of a target identity specified by the latent code $\boldsymbol{c}$ (see Fig.~\ref{fig:conditional_synthsis}). This may also be equivalently viewed as ``transferring'' an expression from one face to another. 
		Unlike in a standard GAN, the synthesized image contains a lot of detail about the target identity due to the incorporation of the identifier $D^2$ and the expression classifier  $D^3$.
		\begin{figure*}[!t]
			\setcounter{figure}{2}
			\centering
			\includegraphics[trim={0 0 0 0.7cm},clip,width=0.75\linewidth]{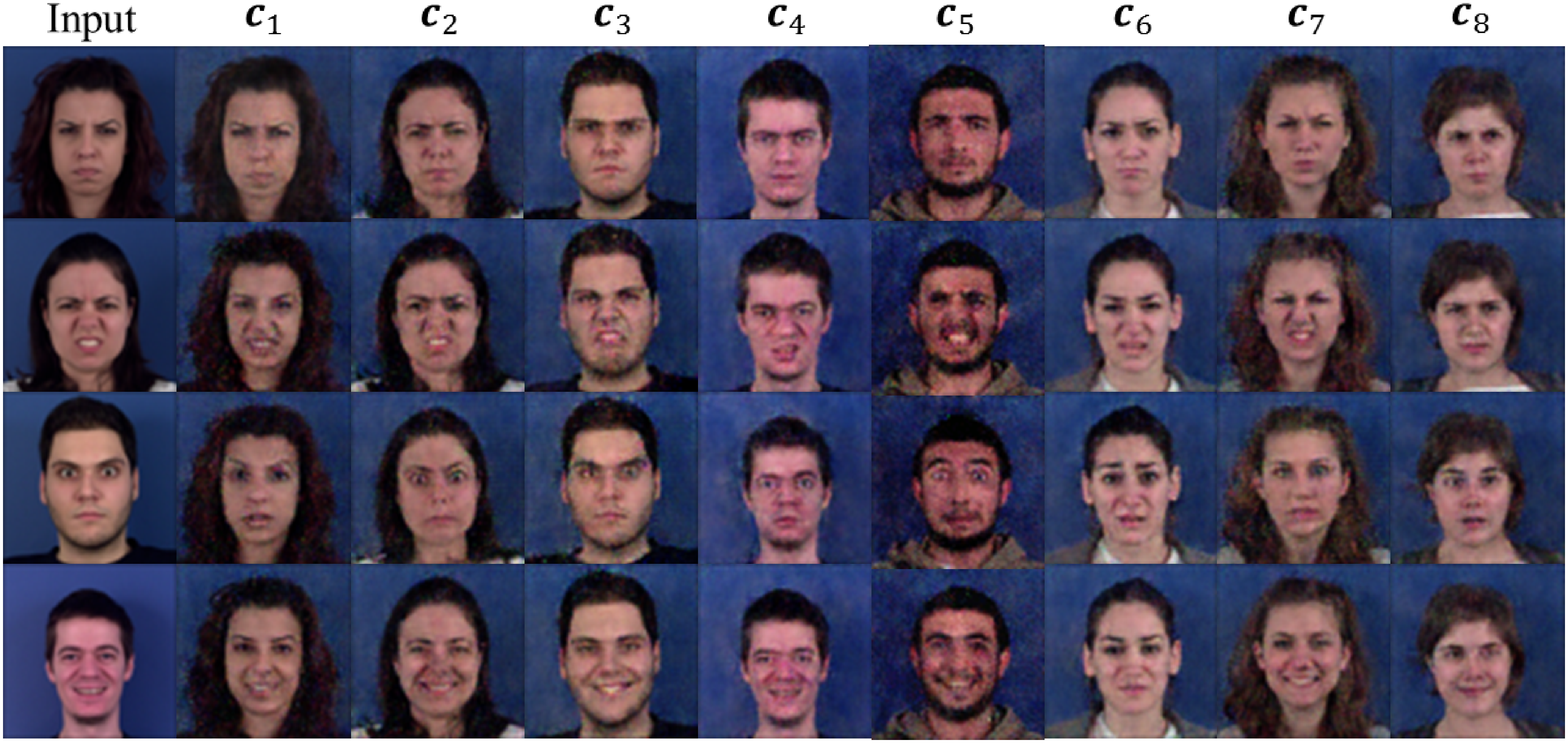}
			\vspace{2mm}
			\includegraphics[width=0.64\linewidth]{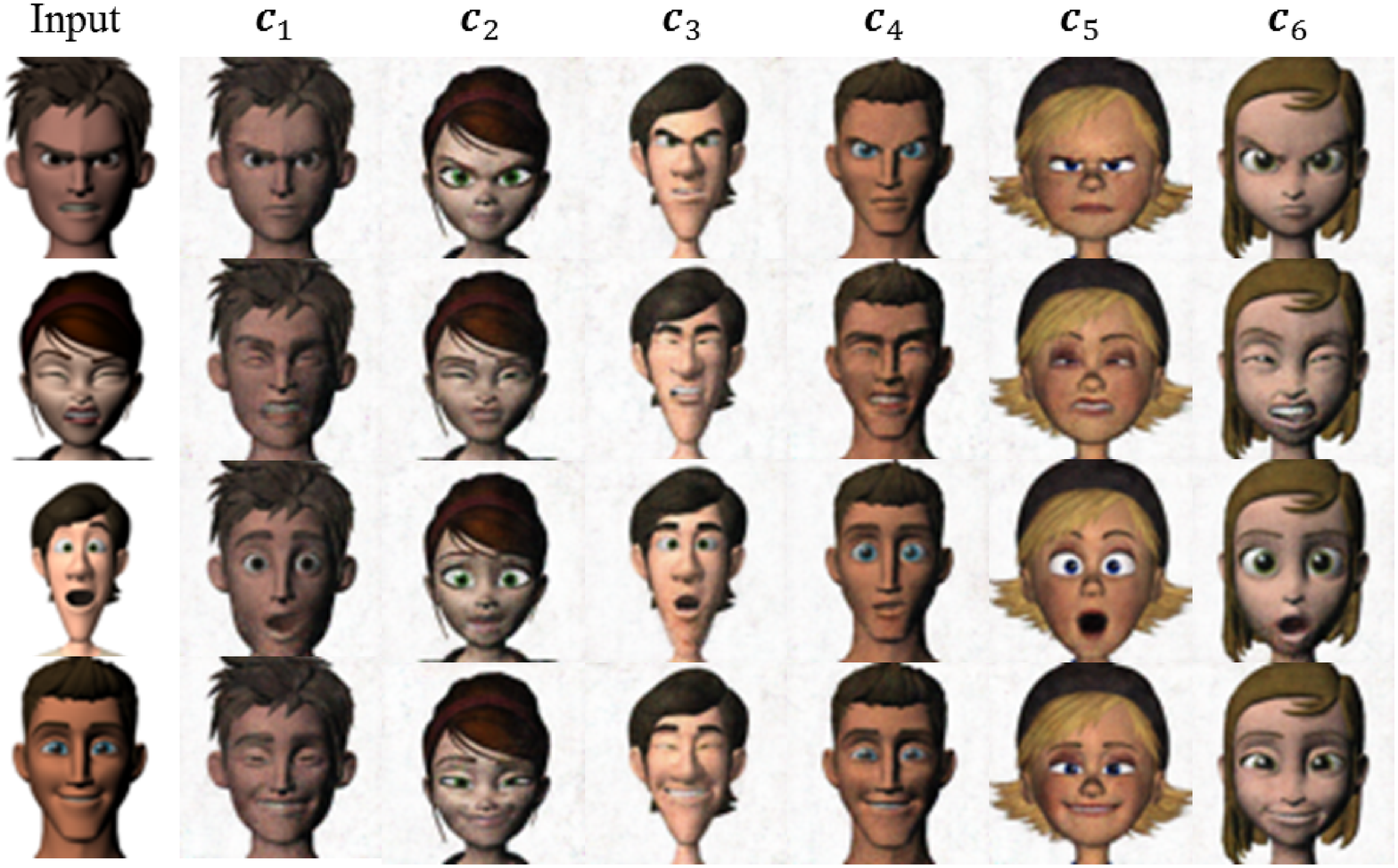}
			\vglue -0cm
			\caption{Examples of identity replacement for both datasets. In each row, from left to right, is an input image followed by synthesized images with identity code $\boldsymbol{c}_i, i = 1,...,N_{id}$.}
			\label{fig:conditional_synthsis}
			\vglue -0.6cm
		\end{figure*}
		
		\subsection{Image Synthesis}
		
		\noindent{\bf Face Image Synthesis without Input Image:}
		Once trained, our model can also synthesize face images without using an input image. 
		This is due to the constraint we impose on the encoder which forces the distribution of the latent representation to follow a prior distribution (in our experiments: $\boldsymbol{f}(\boldsymbol{I})\sim \mathcal{N}(\boldsymbol{0},\boldsymbol{I})$). 
		To generate a new face image, we simply sample a latent vector from the prior distribution and concatenate it with an identity code. Then, we feed the concatenated vector into the decoder for image generation.
		As shown in Fig.~\ref{fig:uncon_syn}, the synthesized images are realistic and the identities are consistent with the identity code $\boldsymbol{c}$.
		While the current model is incapable of controlling the facial expression of a generated image when no input image is given, we believe the synthesized images are useful for other applications, e.g, augmenting the original dataset.
		
		\noindent{\bf Face Image Synthesis for Left-Out Expression:}
		In order to further evaluate the generative capacity of PPRL-VGAN, we conducted experiments where we intentionally left out all samples of a specific facial expression $e$ from subject $i$ in training (images of expression $e$ from other subjects are still used) and then synthesized the left-out expression for subject $i$ after the model had been trained. 
		%
		This was done by feeding the generator $G$ an image with expression $e$ from subject $j$, $j \neq i$, and an identity code $\boldsymbol{c}_i$ with $i$th entry equal to 1 and all other entries 0.
		\begin{figure*}[!t]
			\setcounter{figure}{3}
			\centering{\includegraphics[trim={0 0 0 0.6cm},clip,width=0.7\linewidth]{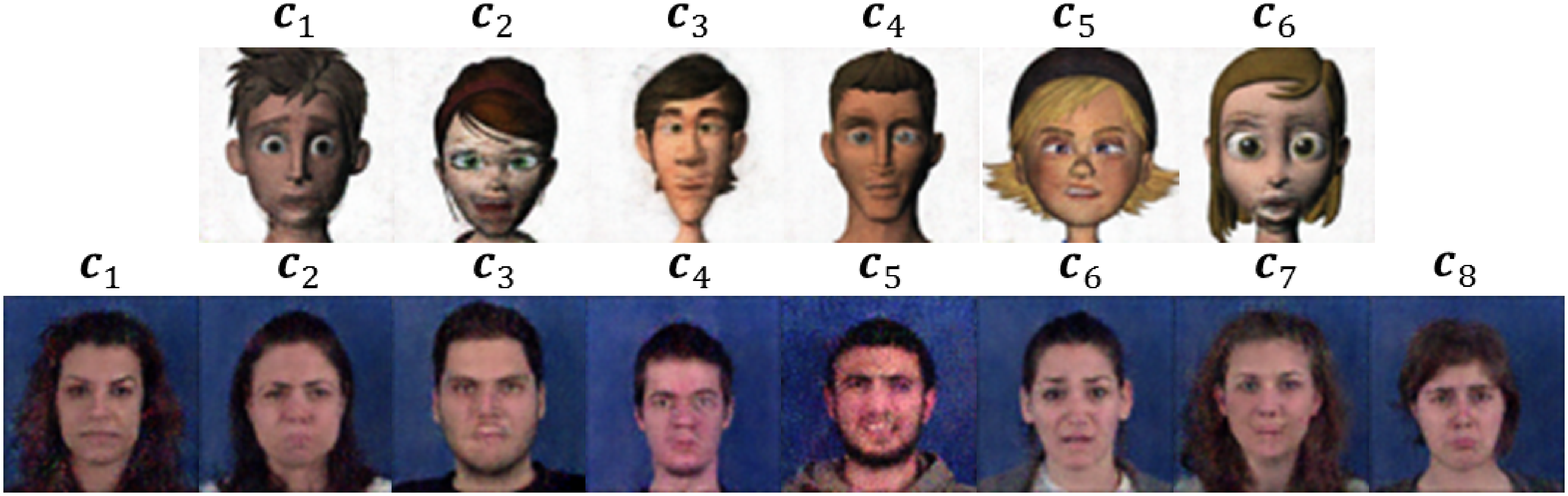}}
			\vglue -0.2cm
			\vspace*{3mm}
			\caption{Image synthesis without input image; $\boldsymbol{f}(\boldsymbol{I}) $ is sampled from $\mathcal{N}(\boldsymbol{0},\boldsymbol{I})$ with identity code $\boldsymbol{c}_i, i = 1,...,N_{id}$.}
			\label{fig:uncon_syn}
			\vglue -0.4cm
		\end{figure*}
		
		%
		%
		Figure~\ref{fig:exp_left} shows examples of left-out  expression synthesis.
		%
		While artifacts are clearly visible, the synthesized images capture the essential traits of a left-out  expression,
		thus validating the generative capacity of PPRL-VGAN. 
		\begin{figure}[!htb]
			\setcounter{figure}{4}
			\centering
			\begin{subfigure}[!htb]{0.18\textwidth}
				\includegraphics[width=1\linewidth]{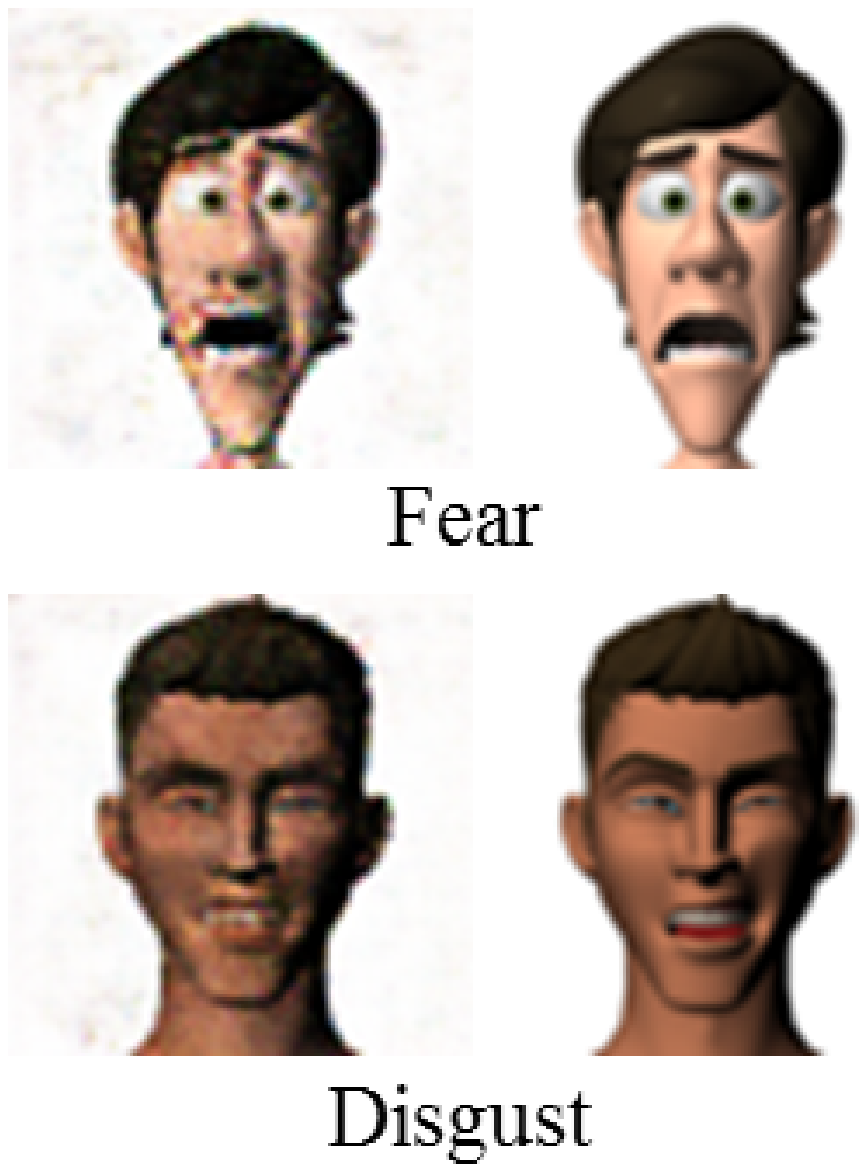}
				\vglue -0.3cm
				\caption{FERG}
				\label{fig:exp_ferg} 
			\end{subfigure}%
			\quad
			\begin{subfigure}[!htb]{0.18\textwidth}
				\includegraphics[width=1\linewidth]{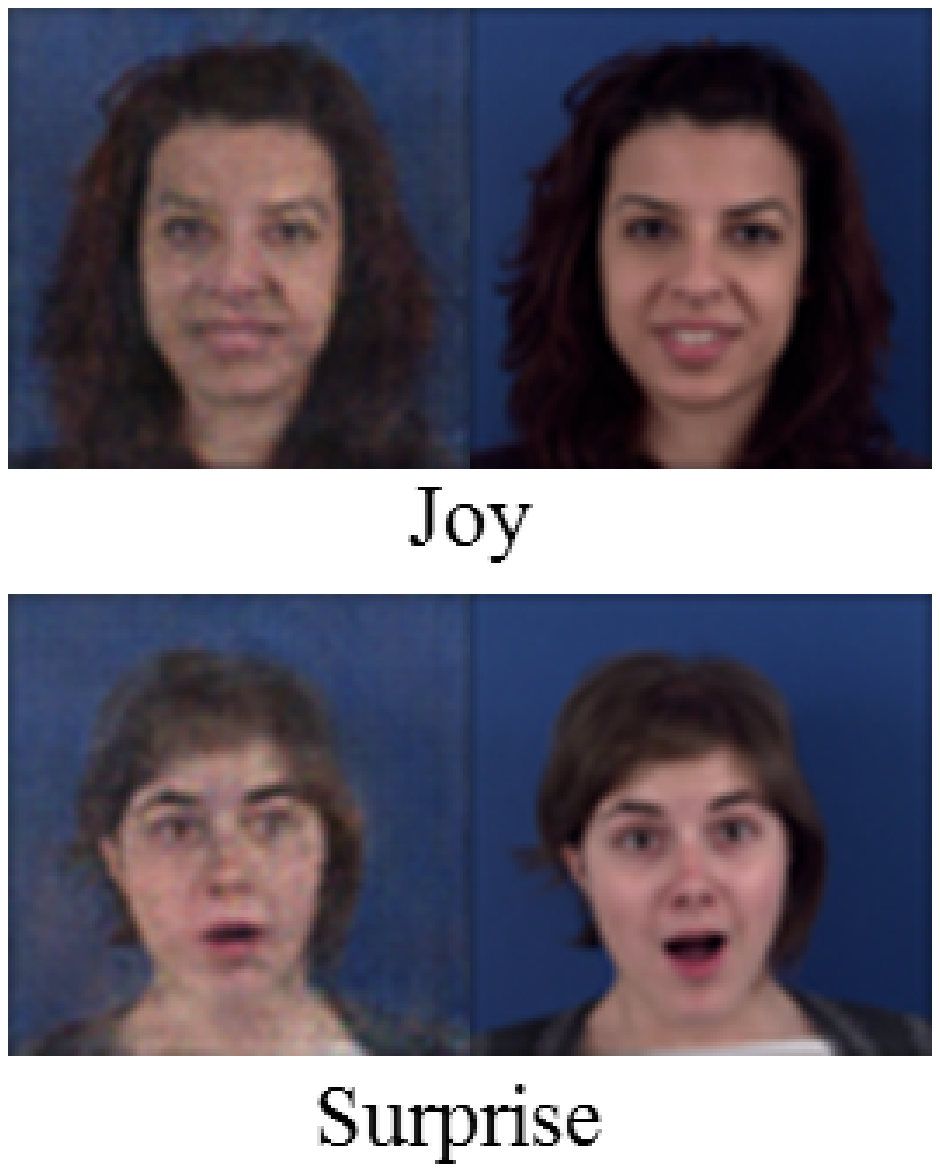}
				\vglue -0.3cm
				\caption{MUG}
				\label{fig:exp_mug}
			\end{subfigure}
			\caption{Image synthesis of left-out expressions (left: synthesized image of a left-out expression; right: corresponding ground-truth image).}
			\label{fig:exp_left}
			\vglue -0.5cm
		\end{figure}
		
		\noindent{\bf Expression Morphing:}
		Facial expression morphing is a challenging problem because a human face is highly non-rigid and significantly deforms across expressions. 
		Most methods perform face morphing in image space. 
		Here, we leverage the latent representation and apply linear interpolation in latent space. 
		Let $\boldsymbol{I}_1$, $\boldsymbol{I}_2$ be  a pair of source images with different expressions for subject $i$
		and $\boldsymbol{f}(\boldsymbol{I}_1)$, $\boldsymbol{f}(\boldsymbol{I}_2)$ their corresponding latent representations. 
		First, we linearly interpolate $\boldsymbol{f}(\boldsymbol{I}_1)$ and $\boldsymbol{f}(\boldsymbol{I}_2)$ in the latent space to obtain a series of new representations $\boldsymbol{f}(\boldsymbol{I}_{interp})$ as follows:
		\begin{equation}
		\boldsymbol{f}(\boldsymbol{I}_{interp}) = (1-\alpha)\boldsymbol{f}(\boldsymbol{I}_1) + \alpha \boldsymbol{f}(\boldsymbol{I}_2), \quad \alpha \in [0,1]
		\end{equation}
		%
		Then, we feed $\boldsymbol{f}(\boldsymbol{I}_{interp})$ and identity code $\boldsymbol{c}_i$ into the decoder to synthesize  images.
		Figure~\ref{fig:morphing} shows two examples of  expression morphing.
		We can see that in both cases, the facial expression changes gradually from left to right. 
		These smooth semantic changes indicate the model is able to capture salient expression characteristics in  $\boldsymbol{f}(\boldsymbol{I})$.
		
		%
		\begin{figure}[!htb]
			\vglue -0.2cm
			\setcounter{figure}{5}
			\centering{\includegraphics[width=1.0\linewidth]{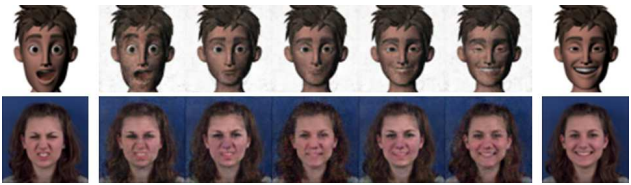}}
			\vglue 0.2cm
			\caption{Examples of expression morphing for FERG (top) and MUG (bottom) datasets. The first and last images in each row are the  source images, while those in-between are  synthesized  by linear interpolation in latent space.}
			\label{fig:morphing}
			\vglue -0.4cm
		\end{figure}

		\noindent{\bf Image completion:}
		PPRL-VGAN can be also applied to an image completion task. 
		We tested two different masks (Fig.~\ref{fig:completion_results}): one covering the eyebrows, eyes and nose, and the other  covering the mouth (each mask occupies $\sim ~7\%$ of the image).
		To complete the missing content of a query image $\boldsymbol{I}_{q}$ of subject $j$, 
		we first pass $\boldsymbol{I}_{q}$ to the encoder to produce a latent representation $\boldsymbol{f}(\boldsymbol{I}_{q})$.
		%
		Then, we feed $\boldsymbol{f}(\boldsymbol{I}_{q})$ and $\boldsymbol{c}_i$ to the decoder for synthesizing a new image  $\boldsymbol{I}^\prime \sim Dec(\boldsymbol{f}(\boldsymbol{I}_{q}), \boldsymbol{c}_i)$.
		Finally, we replace the missing pixel values of $\boldsymbol{I}_{q}$ with values from corresponding locations in $\boldsymbol{I}^\prime$.

		Examples of both successful and unsuccessful image completions are shown in Fig.~\ref{fig:completion_results}.
		%
		Figure~\ref{fig:completion_suc} shows examples for which our model was able to  accurately estimate the missing image content.
		%
		This demonstrates that our model learns correlations between different facial features, for example that opening the  mouth is likely to appear jointly with raising eyebrows.
		However, our model occasionally fails (Fig.~\ref{fig:completion_fail}). 
		One possible reason for this is that some critical facial features (e.g., lowered eyebrows and narrowed eyes in the angry expression) are missing.
		A distortion may also occur when a face in the synthesized images is not accurately aligned with the one in the query image. 
		%
		\begin{figure}
			\setcounter{figure}{6}
			\centering
			\begin{subfigure}[!htb]{0.5\textwidth}
				\includegraphics[width=0.95\linewidth]{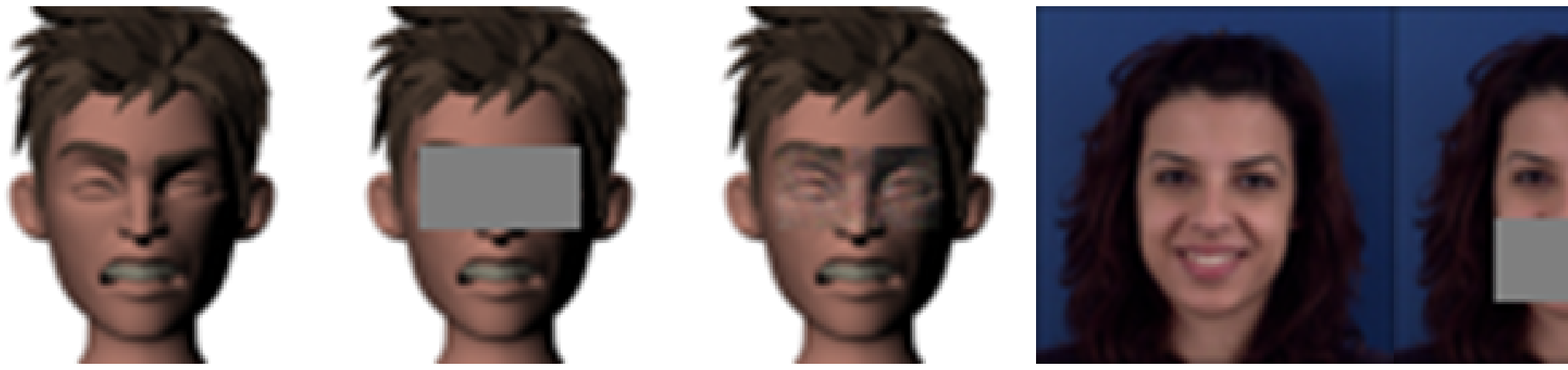}
				\caption{Examples of successful image completion  }
				\label{fig:completion_suc} 
			\end{subfigure}%
			\quad
			\begin{subfigure}[!htb]{0.5\textwidth}
				\includegraphics[width=0.95\linewidth]{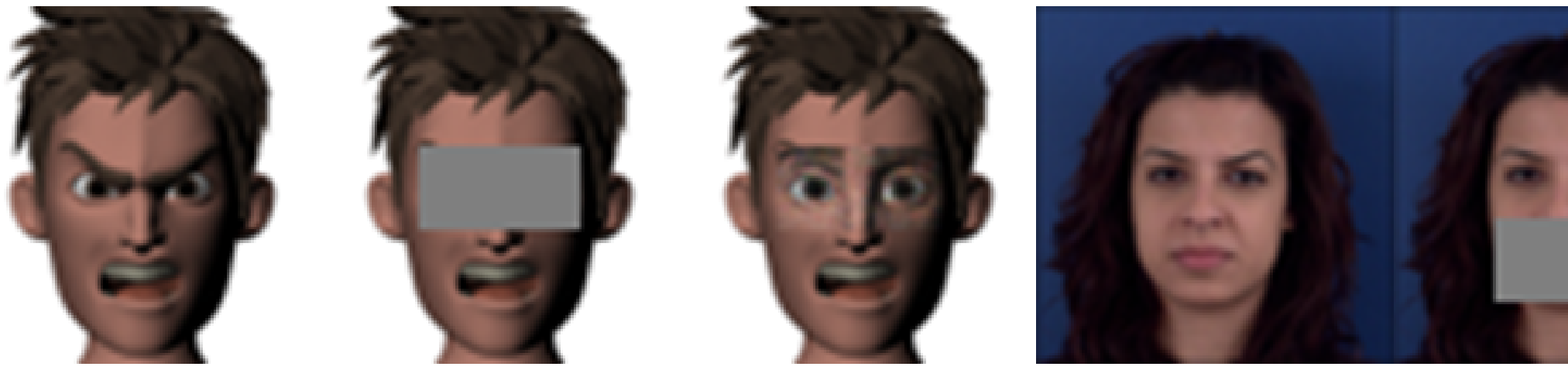}
				\caption{Examples of unsuccessful image completion }
				\label{fig:completion_fail}
			\end{subfigure}
			\caption{Example of image completion for FERG and MUG datasets. From left to right:  original image, masked image and image completion result. Note that the original images are excluded from the training set.}
			\label{fig:completion_results}
			\vglue -0.6cm
		\end{figure}
		\section{Conclusion}
		We presented a PPRL-VGAN for privacy-preserving facial expression recognition and face image synthesis. 
		We proposed a novel architecture combining a VAE and a GAN to create an identity-invariant representation of a face image that also permits synthesis of an expression-preserving and realistic version.
		%
		Experimental results on two public facial expression datasets demonstrate that our approach strikes a balance between privacy preservation and data utility. 
		In addition, the proposed model can support a variety of applications like expression morphing and image completion. 
		Generalizing the proposed framework to handle input images from unseen persons is part of our ongoing research.
\section{Acknowledgement}
We gratefully acknowledge the support of NVIDIA Corporation with the donation of the Titan X Pascal GPU used for this research.

{\small
\bibliographystyle{ieee}
\bibliography{egbib}
}
\end{document}